\documentclass[review]{elsarticle}
\def\allfiles{}
\usepackage{lineno,hyperref}
\modulolinenumbers[5]

\journal{Journal of \LaTeX\ Templates}

\usepackage{mystyle}

\biboptions{authoryear}

\begin{document}

\begin{frontmatter}

\title{A Biologically Plausible Learning Rule for Perceptual Systems of organisms that Maximize Mutual Information}
% \tnotetext[mytitlenote]{Fully documented templates are available in the elsarticle package on \href{http://www.ctan.org/tex-archive/macros/latex/contrib/elsarticle}{CTAN}.}

% \address{Radarweg 29, Amsterdam}
% \fntext[myfootnote]{Since 1880.}

% \author[mymainaddress,mysecondaryaddress]{Elsevier Inc}

\author{Tao LIU\corref{mycorrespondingauthor}}
\cortext[mycorrespondingauthor]{Corresponding author}
\ead{ntobby.liu@gmail.com}

\address{Yantai, Shandong, China}
% \address[mysecondaryaddress]{360 Park Avenue South, New York}

\begin{abstract}
It is widely believed that the perceptual system of an organism is optimized for the properties of the environment to which it is exposed.
A specific instance of this principle known as the Infomax principle holds that the purpose of early perceptual processing is to maximize the mutual information between the neural coding and the incoming sensory signal.
In this article, we present a method to implement this principle accurately with a local, spike-based, and continuous-time learning rule.
\end{abstract}


\end{frontmatter}

% \linenumbers
\ifx\allfiles\undefined
\usepackage{my}
\begin{document}
\fi

\section{Introduction}\label{sc:1}
Consider a neural perceptual system being exposed to an external environment.
The system has certain internal state to represent external events.
There is strong behavioral and neural evidence  \citep[e.g.,][]{ernst2002humans,gabbiani1998principles} 
that the internal representation is intrinsically probabilistic \citep{knill2004bayesian}, in line with the statistical properties of the environment.

We mark the input signal as $x$. The perceptual representation would be a probability distribution conditional on $x$, denoted as $p(y\mid x)$.
According to the Infomax principle \citep{attneave1954some,barlow1961possible,linsker1988self}, the mission of the perceptual system is to maximize the mutual information (MI) between the input (sensory signal) $x$ and the output (neuronal response) $y$, which can be written as
\begin{equation}\label{eq1.00}
    \max_{p(y\mid x)}I(x;y),
\end{equation}
where $I(x;y)\equiv H(y)-H(y\mid x)$ is the MI, $H(y)\equiv -\E_{y \sim p(y)} \log p(y)$ and $H(y\mid x)\equiv -\E_{x,y \sim p(x,y)} \log p(y\mid x)$ are marginal and conditional entropies\footnote{In this article, the unit of information entropy is nat, and $\log$ is based on $e$.} respectively, and $\E$ represents the expection.

% The first achievement of this article is to find an accurate solution for \eqref{eq1}.
Despite the simplicity of the statement, the MI is generally intractable for almost all but special cases.
In the past, to work around this difficulty, a variety of approximate methods were considered, including to optimize the Fisher `Information' \citep{brunel1998mutual}, the Variational Information Maximization \citep{agakov2004algorithm}, and other approaches relying on approximate gaussianity of data distribution \citep{linsker1992local} or zero/low-noise limit \citep{bell1995information,nadal1994nonlinear}, etc.
% It seems that certain approximation is inevitable when the analytical calculation of the MI is not available.
However, an accurate\footnote{The word ``accurate" used in this article does not suggest a closed form, but that the result can be made as accurate as you like.} solution is always prioritized, for it may bring great evolutionary advantages over approximate ones. So in this article, we first propose a novel gradient-based method to optimize the MI accurately. This method avoids the limitation that gradient method is only applicable to differentiable functions by introducing an auxiliary distribution.
% To the best of my knowledge, this is the first accurate solution when both $p(y)$ and $p(y\mid x)$ are parametric.

In physical neural systems, the calculation of $p(y\mid x)$ will be executed by neural circuits, and in artificial systems, it will be executed by artificial components.
In both cases, the capacity constraint of the system needs to be considered \citep{barlow1961possible}.
Due to this constraint, only latent variables in the environment deserve to be represented, while other information, such as noise, does not.
The ``latent variables" here refer to the invariant or slowly varying features \citep{wiskott2002slow} underlying the sensory input, such as the location of an object \citep{battaglia2003bayesian}, the direction of random dot motion \citep{newsome1989neuronal, britten1992analysis}, the orientation of the drifting grating stimulus \citep{berkes2005slow}, etc.
A general approach for estimating latent variables online from an input stream is Bayesian filtering \citep{chen2003bayesian}, and in this article, I will demonstrate that it can work well with the Infomax principle, that is, Bayesian filtering specifies an inductive bias for calculating the perceptual representation, and the Infomax principle guides the optimization of that calculation.

The final and most important contribution of this article is to present a biologically plausible learning algorithm based on the Infomax principle and the Bayesian filtering approach mentioned above.
Biological neural systems distinguish themselves from popular artificial ones by their locality of operations, spike-based neural coding, and continuous-time dynamics. 
Among these properties, the locality of operations is the hardest to achieve, for it restricts the learning rule to only involve variables that are available locally in both space and time.
A large body of research, including \citet{foldiak1989adaptive}, \citet{rubner1989self},  \citet{Krotov7723}, etc., has used local learning rules motivated by Hebb’s idea, but these rules are postulated rather than derived from a principled cost function.
In this article, I demonstrate that the property of spacial locality can be achieved under the mean-field approximation, and the property of temporal locality can be achieved under certain assumptions of the chronological order of sensory stimuli.
The other two biological properties will also be achieved by using an inhomogeneous Poisson model of spike generation.
% Other properties of biological plausibility are also met in my work as well. 

{
}
The rest of the article is organized as follows.
Section~\ref{sc:2} introduces the unbiased algorithm to maximize the MI accurately, 
Section~\ref{sc:3} introduces the mean-field approximation to make the algorithm to be spatially local, 
Section~\ref{sc:4} introduces the Bayesian filtering technique to make the algorithm to be temporally local, and 
Section~\ref{sc:5} introduces the inhomogeneous Poisson process to achieve the properties of pike-based neural coding and continuous-time dynamics.
The techniques developed in this article are incremental, that is, in Section~\ref{sc:5}, we will obtain a learning algorithm with all of the above properties.

\ifx\allfiles\undefined
\bibliography{reference}
\end{document}
\fi
\ifx\allfiles\undefined
\usepackage{my}
\begin{document}
\fi

\section{The accurate method to maximize the MI}
\label{sc:2}
In mathematics, the optimization problem \eqref{eq1.00} can be solved with a gradient ascent algorithm.
Since there is no eligible notation for gradient ascent or descent algorithms yet, we set up one here.
The symbols 
\begin{equation*}{
   \asc_{\delta f} \delta J[f] \text{, and } \des_{\delta f} \delta J[f]}
\end{equation*}
will be used to denote the gradient ascent and descent algorithms for solving
\begin{equation*}
    \max_{f}J[f] \text{, and }\displaystyle \min_{f}J[f],
\end{equation*}
respectively, in which
$f$ is a function, $J[f]$ is a functional of $f$, $\delta f$ is a small change in $f$, and $\delta J[f]$ is the variation\footnote{The variation of a functional is the analogous concept to the differential of a function.} of $J[f]$ due to $\delta f$. 
With this notation, the gradient ascent algorithm for the  problem \eqref{eq1.00} can be denoted as \begin{align}\label{eq2.00}
    \asc_{\delta p(y\mid x)}\delta I(x;y),
\end{align}
where $\delta I(x;y)$ is the variation of $I(x;y)$ due to $\delta p(y\mid x)$.
% The distribution $p(x)$ is fixed but unknown, and the joint distribution $p(x,y)$ is determined by 
% The Infomax principle can be rewritten from \eqref{eq1.00} into
% \begin{equation*}\label{eq2}
%     \max_{p(y\mid x)} \{ H(y)-H(y\mid x) \}.
% \end{equation*}
% Because $p(x)$ is unknown, we can only apply Monte Carlo method to tackle a problem involving it.
% For example, $\displaystyle H(y\mid x)=\E_{x\sim p(x)}\int p(y\mid x)\log p(y\mid x)\diff y$
% The difficulty in computing the MI lies in the computation of $H(y)$, because $p(x)$ is unknown.
% However, our goal is to maximize the MI, but not to compute it.
% More specifically,

In the real environment, the input signal $x$ is sampled from an unknown but fixed distribution $p(x)$.
Any small change in $p(y\mid x)$ will lead to a corresponding variation of $I(x;y)$. 
We can derive from formulas of information theory that \emph{(Proof.\ref{app:1})}
\begin{equation}\label{eq2.10}
    \delta I(x;y) = \E_{x\sim p(x)}\int \log(\frac{p(y\mid x)}{p(y)})  \delta p(y\mid x)\diff y,
\end{equation}
where $\delta p(y\mid x)$  represents the small change in $p(y\mid x)$, and $\delta I(x;y)$ represents the variation of $I(x;y)$ due to $\delta p(y\mid x)$. 

The marginal distribution $p(y)$ in \eqref{eq2.10} cannot be computed directly, but we can create an auxiliary distribution $q(y)$ that is equal to it.
Since $p(y)$ may change as $p(y\mid x)$ changes with optimization, the auxiliary $q(y)$ should also be trained continuously to keep pace with $p(y)$. The algorithm used for training $q(y)$ is \emph{(Proof.\ref{app:1.2})}
\begin{equation}\label{eq2.20}
    \des_{\delta q(y)}\E_{x\sim p(x)} \int \left(q(y) - p(y\mid x)\right)\delta q(y)  \diff y.
\end{equation}

Now, we obtain a gradient-based method to maximize the MI accurately, which is composed of \eqref{eq2.20} and 
\begin{align}
    \asc_{\delta p(y\mid x)}\E_{x\sim p(x)}\int \log(\frac{p(y\mid x)}{q(y)})  \delta p(y\mid x)\diff y.\label{eq2.30}
\end{align}
% We can name the algorithm in \eqref{eq2.20} as the auxiliary process, and the algorithm in \eqref{eq2.30} as the main process.
In principle, the algorithm in \eqref{eq2.30} is executed only when the the algorithm in \eqref{eq2.20} reaches a stable state.
In practice, this can be done by setting two different learning rates for the two processes, respectively.

\paragraph{Modeling}
If $p(y)$ and $p(y\mid x)$ are parametric, we can model them with artificial models, such as ANNs (artificial neural networks), and then execute \eqref{eq2.20} and \eqref{eq2.30} by sampling $x$ repeatedly (that is, the Monte Carlo).
As this method is unbiased, the result can be made as accurate as you like.
In contrast, if $p(y)$ or $p(y\mid x)$ is not parametric, in addition to sampling $x$, it is also needed to sample or grid $y$ to integrate over it numerically, which is theoretically feasible, but extremely time consuming in practice.

\paragraph{Chase game}One interesting finding about \eqref{eq2.20} and \eqref{eq2.30} is that they are playing a chase game. Given that 
\begin{equation}
    \begin{cases}
        q(y) - p(y\mid x)>0 \text{, }\log(\frac{p(y\mid x)}{q(y)}) <0 & \text{ if } q(y) > p(y\mid x) \\
        q(y) - p(y\mid x)<0 \text{, }\log(\frac{p(y\mid x)}{q(y)}) >0& \text{ if } q(y) < p(y\mid x),
    \end{cases}
\end{equation}
the descent algorithm in \eqref{eq2.20} always makes $q(y)$ close to $p(y\mid x)$, and the ascent algorithm in \eqref{eq2.30} always makes $p(y\mid x)$ away from $q(y)$.
\citet{schmidhuber1992learning} has used a similar approach, but the formulas used in that work are postulated rather than derived from a certain principle.

{
}
\ifx\allfiles\undefined
\bibliography{reference}
\end{document}
\fi
\ifx\allfiles\undefined
\usepackage{my}
\begin{document}
\fi

\section{The mean-field approximation and spatial locality}
\label{sc:3}
Now, we consider the neuronal response $y\equiv (y_1y_2\cdots y_n)$ as an $n$-tuple vector, in which $y_i$ represents the response of a single neuron (or perhaps a cortical column).
The conditional probability of $y_i$ is denoted by $p(y_i\mid xy_{-i})$, where $y_{-i}$ is the abbreviation for $(y_1\cdots y_{i-1}y_{i+1}\cdots y_n)$.
Our goal is to find a learning algorithm to optimize the calculation of $p(y_i\mid xy_{-i})$ in a spatially local manner.

Let us imagine a nut with a kernel and a shell. 
The kernel is the composite of the representation neuron numbered $i$ and some other auxiliary components. It is responsible for calculating and optimizing $p(y_i\mid xy_{-i})$ every time the inputs $x$ and $y_{-i}$ are inputted, and then it outputs $y_i$. 
The calculation capacity of it is limited, that is, it can only calculate the (conditional) probability of a scalar variable, such as $p(y_i\mid xy_{-i})$, $p(y_i\mid x)$, and $p(y_i\mid y_{-i})$, but not $p(y_{-i}\mid x)$ because the $y_{-i}$ and $x$ can both be vectors.
% generates the probabilistic response $y_i$ from the inputs $x$ and $y_{-i}$.
% The computations and optimizations are run by the kernel.
% The probability of $y_i$ is computed through some mechanism inside the kernel.
The shell of the nut is a metaphor for the property of spatial locality. It keeps all signals except $x$, $y_{-i}$, and $y_i$ from passing through it.
% In the nut, the probability $p(y_i\mid xy_{-i})$ is calculated and optimized every time $x$ and $y_{-i}$ are given.

The optimization of $p(y_i\mid xy_{-i})$ each time may bring a small change $\delta p(y_i\mid xy_{-i})$.
Due to the constraints introduced in the last paragraph, the small change $\delta p(y_i\mid xy_{-i})$ should not depend on the distribution $p(y_{-i}\mid x)$, which is incalculable, nor should it bring any changes to other probabilities $p(y_j\mid xy_{-j})\mid_{j\ne i}$. These two requirements are met if and only if \emph{(see Proof.\ref{app:300})}
\begin{align}\label{eq3.10}
    p(y\mid x)=\prod_{i=1}^{n} p(y_i\mid x),
\end{align}
which was known as the mean-field approximation \citep{blei2017variational}. 
% It is also one of the most commonly used approximations in variational inference.
In computational neuroscience, \eqref{eq3.10} is the formalization of the independent-neuron hypothesis \citep[see][chapter 1.5]{dayan2001theoretical}. This hypothesis generally states that individual neurons act independently, which is a useful simplification and is not in gross contradiction with experimental data.

% , and the variation $\delta p(y_i\mid xy_{-i})$ degenerates into $\delta p(y_i\mid x)$, correspondingly
% According to \eqref{eq3.00},
% \begin{align}\label{eq3.20}
%       \delta I(x;y)  = \sum_{i=1}^{n}\E_{\begin{subarray}{l} x\sim p(x)\\y_{-i}\sim p(y_{-i}\mid x)\end{subarray}}\int \log\frac{p(y_i\mid x)}{p(y_i\mid y_{-i})} \delta p(y_i\mid x) \diff y_i. 
% \end{align}
Under the mean-field approximation, the conditional probability for $y_i$ will degenerate from $p(y_i\mid xy_{-i})$ into $p(y_i\mid x)$, and we will also have
\begin{align}\label{eq3.20}
    \delta p(y\mid x)=\sum_{i=1}^{n}p(y_{-i}\mid x)\delta p(y_i\mid x),
\end{align}
and \emph{(see Proof.\ref{app:310})}
\begin{align}\label{eq3.30}
    \delta I(x;y)=\sum_{i=1}^{n}\E_{\begin{subarray}{l} x\sim p(x)\\y_{-i}\sim p(y_{-i}\mid x)\end{subarray}}\int \log\frac{p(y_i\mid x)}{p(y_i\mid y_{-i})} \delta p(y_i\mid x) \diff y_i.
\end{align}
% When $\displaystyle p(y\mid x)=\prod_{i=1}^{n} p(y_i\mid x)$, the conditional distribution $p(y_i\mid xy_{-i})$ in \eqref{eq3.00} will degenerate into $p(y_i\mid x)$. Besides, the training signal of the factor $p(y_i\mid x)$ does not propagate into other factors of $p(y\mid x)$, 
According to \eqref{eq3.20} and \eqref{eq3.30}, the optimization problem \eqref{eq1.00} now can be solved with a distributed algorithm with the $i$'th part of it as \emph{(see Proof.\ref{app:3.5})}
\begin{align}\label{eq3.35}
 \asc_{\phantom{_{-}}\delta p(y_i\mid x)\phantom{_{i}}}\E_{\begin{subarray}{l} x\sim p(x)\\y_{-i}\sim p(y_{-i}\mid x)\end{subarray}}\int \log\frac{p(y_i\mid x)}{p(y_i\mid y_{-i})} \delta p(y_i\mid x) \diff y_i.   
\end{align}
% Each part of the method is responsible for the training of one factor of $p(y\mid x)$.
Similar to the algorithms \eqref{eq2.20} and \eqref{eq2.30} derived in the last section, \eqref{eq3.35} is solved by the chase of two algorithms \eqref{eq3.40} \emph{(see Proof.\ref{app:3})} and \eqref{eq3.50}.
\begin{empheq}[left={\empheqlbrace}]{alignat=2}
    &\des_{\delta q(y_i\mid y_{-i})}&&\E_{\begin{subarray}{l} x\sim p(x)\\y_{-i}\sim p(y_{-i}\mid x)\end{subarray}} \int \left(q(y_i\mid y_{-i}) - p(y_i\mid x)\right)\delta q(y_i\mid y_{-i})  \diff y_i
    \label{eq3.40}\\
    &\asc_{\phantom{_{-}}\delta p(y_i\mid x)\phantom{_{i}}}&&\E_{\begin{subarray}{l} x\sim p(x)\\y_{-i}\sim p(y_{-i}\mid x)\end{subarray}}\int \log\frac{p(y_i\mid x)}{q(y_i\mid y_{-i})} \delta p(y_i\mid x) \diff y_i.\label{eq3.50}
\end{empheq}
In principle, the algorithm in \eqref{eq3.50} is executed only when the algorithm in \eqref{eq3.40} is at a stable point.

\paragraph{The scale of spatial locality}
So far, we have realized the property of spatial locality on the nut scale, but we have not yet resolved it for the learning algorithms executed inside the shell. 
Since our learning algorithm is gradient-based, it is spatially local only when both $q(y_i\mid y_{-i})$ and $p(y_i\mid x)$ can be modeled with no hidden units.
% for the gradient-based algorithms used to train $q(y_i\mid y_{-i})$ and $p(y_i\mid x)$.
% Since the learning algorithms of $q(y_i\mid y_{-i})$ and $p(y_i\mid x)$ are gradient-based, they are local 
% % the spatial locality of the algorithms inside the nut for training $q(y_i\mid y_{-i})$ and $p(y_i\mid x)$.
% If we use ANNs to model them, the learning method of them is local only if the ANNs are single layered.
% Since only single-layer ANNs can be trained with local learning ruels, if we use ANNs model $q(y_i\mid y_{-i})$ and $p(y_i\mid x)$, they have to be single-layer.
% If we can use ANNs to model $q(y_i\mid y_{-i})$ and $p(y_i\mid x)$, the gradient algorithm is local only when these ANNs are single-layer.
This will be further discussed in the end of Section \ref{sc:5}.

% We have resolved the property of spatial locality beyond the nut scale through a distributed algorithm.
% Now we turn to the algorithms preformed inside the nut.
% We can use ANNs to model $q(y_i\mid y_{-i})$ and $p(y_i\mid x)$.
% Since we have resolved the property of spatial locality beyond the scale of the nut, the remaining question is whether $q(y_i\mid y_{-i})$ and $p(y_i\mid x)$, which are calculated inside the nut, can be modeled with single-layer neural networks, or must they be modeled with deep neural networks.
% This will be discussed in the next section.

{
}

\ifx\allfiles\undefined
\bibliography{reference}
\end{document}
\fi
\ifx\allfiles\undefined
\usepackage{my}
\begin{document}
\fi
\section{Bayesian filtering and temporal locality}
\label{sc:4}
Now we introduce time into perception by proposing the concept of perceptual events. 
Perceptual event relates to the generation of a stable and informative perceptual representation through the accumulation of sensory stimuli over a short period of time. It is the smallest granularity in conscious perception, that is, all stimuli during this period would be grouped and subjectively interpreted as a single event, and so their chronological order cannot be judged by the agent \citep{vanrullen2003perception}.
% Consciousness is unable to distinguish the temporal order of sensory stimuli within an event. in an event
% Perceptual event refers to the process in which the perceptual system accumulates sensory stimuli.
% from the first sensory stimulus received by the perceptual system to the final perceptual representation entering consciousness by accumulating stimuli.
For a perceptual event, the input $x$ is the time series
\begin{equation*}
x_{1:T}\equiv(x_1x_2\cdots x_T),   
\end{equation*}  where 
$x_t\mid_{t\in (1:T)}$ represents the sensory stimulus at time step $t$, 
$T$ is the total time steps of the period of accumulation, and the size of time step is $\Delta t$.
The generated perceptual representation is $p(y\mid x_{1:T})$.
% , and the resulting perceptual representation to be fed into consciousness is $p(y\mid x_{1:T})$.
% A virtual downstream conscious system works at the granularity of events, and it cannot distinguish (or is simply not interested in) the temporal order of the stimuli within the event.
%在之前，经历一段演化过程。
% The perceptual perception is the state of a perceptual system.

As we discussed in the section \ref{sc:1}, the representation $p(y\mid x_{1:T})$ can be interpreted as the estimate of certain latent variables according to the observation $x_{1:T}$.
% Our goal is to find temporally local methods to compute and train $p(y\mid x_{1:T})$.
The perceptual system  goes through a series of transient states before obtaining it. 
Naturally, these transient states can be interpreted as the estimate of those variables based on the incomplete observations at the time.
To be more specific, the transient state at time step $t$ is the conditional distribution $p(y\mid x_{1:t})$, and the spontaneous state before any stimuli arriving is the prior distribution $p(y)$.
% The latent variables encoded by $y$ can be considered constant during the event.

The property of temporal locality indicates that in each time step $t$, the new state $p(y\mid x_{1:t})$ is calculated only according to the current stimulus $x_t$ and the old state $p(y\mid x_{1:t-1})$. 
In Bayesian theory,  
\begin{equation}
    \label{eq:400}
    \begin{aligned}
    p(y\mid x_{1:t}) \propto p(y\mid x_{1:t-1})p(x_t\mid y),\\
    \end{aligned}
\end{equation}
where $p(y\mid x_{1:t-1})$ is the prior probability, and $p(x_t\mid y)$ is the likelihood probability.
Therefore, $p(y\mid x_{1:T})$ can be obtained recursively through
\begin{equation}
    \label{eq:420}
    \begin{aligned}
    p(y\mid x_{1:T}) \propto p(y)p(x_1\mid y)p(x_2\mid y)\cdots p(x_T\mid y).
    \end{aligned}
\end{equation}
Since $p(y\mid x_{1:T})$ does not encode the temporal information of a single stimulus, the likelihood $p(x_t\mid y)|_{t\in (1:T)}$ is assumed to be time-invariant, that is, \begin{equation}\label{eq:430}
    p(x_{1}\mid y)\overset{d}{=}  p(x_{2}\mid y)\overset{d}{=}  \cdots  p(x_{T}\mid y),
\end{equation}
where the symbol $\overset{d}{=}$ denotes equality in distribution.
Under this assumption, the expression \eqref{eq:420} is actually the formula of Bayesian filtering \citep{sarkka2013bayesian}, except that in the standard definition of Bayesian filtering, the latent variables can change according to a certain transition probability.

The goal of the perceptual system in learning is to maximize the MI between $x_{1:T}$ and $y$:
\begin{equation}\label{eq:435}
    \max_{p(y\mid x_{1:T})}I(x_{1:T};y)
\end{equation}
Constrained by the property of temporal locality, the gradient information that guides learning is only available for the calculation in the last time step, so only $p(x_T\mid y)$ can be involved in optimization, while $p(y)$ and $p(x_t\mid y)|_{t\in (1:T-1)}$ can not.
% 
% This is shallow learning in which only $p(x_T\mid y)$ is trained, while deep learning, on the contrary, trains all $p(x_t\mid y)|_{t\in (1:T)}$ and $p(y)$ together.
% this is a shallow learning where only the calculation of $p(x_T\mid y)$ to be trained, in contrast with the deep learning where all calculations of $p(x_t\mid y)|_{t\in (1:n)}$ are trained together.
However, if we assume that the probability $p(x_{1:T})$ is invariant for any permutation in $x_{1:T}$, then the shallow training on $p(x_T\mid y)$ can deliver the same results as the deep training performed on all likelihood functions simultaneously. 
To use an example, if the observations $(x_1x_2x_3)$, $(x_2x_3x_1)$, and $(x_3x_1x_2)$ are equally likely to occur, then the training of the last likelihood probability in each event is equivalent to the training of all involved likelihood probabilities in one event.
% As for the prior distribution $p(y)$, it can be directly set to be the maximum entropy distribution.
% We will not formalize the learning algorithm as what we did in \eqref{eq2.20} and \eqref{eq2.30}, because the dependence of $\delta p(y\mid x_{1:T})$ on $\delta p(x_T\mid y)$ is complex.
% However, in the next section, we will introduce a very neat form.

By introducing time $t$ into \eqref{eq3.10}, we have that for all $t\in (1:T)$,
\begin{equation}\label{eq:440}
    p(y\mid x_{1:t})=\prod_{i=1}^{n} p(y_i\mid x_{1:t}).
\end{equation}
From \eqref{eq:420}, \eqref{eq:430}, and \eqref{eq:440}, we can prove that \emph{(Proof.\ref{app:410})} for all $i\in (1:n)$,
\begin{equation}\label{eq:450}
   p(y_i\mid x_{1:T}) \propto p(y_i)p(x_1=x_1\mid y_i)p(x_1=x_2\mid y_i)\cdots p(x_1=x_T\mid y_i)
\end{equation}
% Similar to \eqref{eq3.10}, the expression \eqref{eq:440} allows the learning algorithm to be spatially local, and similar to \eqref{eq:420} and \eqref{eq:430}, the expressions 
% \eqref{eq:440} and \eqref{eq:450} allow the learning algorithm to be temporally local.
The expression \eqref{eq:440}, like  \eqref{eq3.10}, allows us to develop a learning algorithm with the property of spatial locality, and the expression \eqref{eq:450}, as discussed in the last paragraph, allows us to make this algorithm to be temporally local.
However, we will not formalize this algorithm as what we did in the last section, because the dependence of $\delta p(y_i\mid x_{1:T})$ on $\delta p(x_1=x_T\mid y_i)$ is very verbose.

% \eqref{eq3.10}, and the expressions in \label{eq:450} is parallel to \eqref{eq:420} and \eqref{eq:430}.
% The formulas \eqref{eq:440} \eqref{eq:450} can be seen as the heterozygote of \eqref{eq3.10} and , can lead to a learning algorithm with spatial-temporal locality.

\paragraph{Temporal information and consciousness}
We have used two assumptions in this section, one is that the distribution $p(x_t\mid y)$ is time-invariant, and the other is that the probability $p(x_{1:T})$ is invariant for permutations in $x_{1:T}$.
Both of them relate to the idea that the chronological order in $x_{1:T}$ conveys no or little perceptible information, which is generally true only for a very short period of time, like 20--\SI{50}{\ms} \citep{kristofferson1967successiveness,hirsh1961perceived}.
We think that the extraction of temporal information from a sensory stream on a longer time scale should be attributed to the function of advanced consciousness rather than perception, whereas the topic of consciousness will not be covered in this article. 
\paragraph{Short-term plasticity}
Although the property of temporal locality limits that the new state can only be calculated according to the current stimulus and the last state, it does not prohibit the system itself from being affected by experience short-termly in the order of hundreds or thousands of milliseconds.
For example, the short-term plasticity can enable the system to respond similarly to a similar stimulus received not long ago\citep{fischer2014serial}.
However, this topic will not be covered in this article either.

{
}

\ifx\allfiles\undefined
\bibliography{reference}
\end{document}
\fi
\ifx\allfiles\undefined
\usepackage{my}
\begin{document}
\fi

\section{Poisson model of continuous-time spike generation}\label{sc:5}
In physical neural systems, the response of a neuron evoked by an input is a sequence of spikes.
A complete description of the stochastic relationship between the input and the response would require us to know the probabilities corresponding to every possible sequence of spikes. 
However, the number of the sequences is typically so large that it is impossible to determine or even roughly estimate all of their probabilities of occurrence. 
Instead, we must rely on some statistical model.
In this work, we assume that the spikes of a single neuron are statistically independent, which is referred to as the independent spike hypothesis \citep[see][]{rieke1999spikes,heeger2000poisson,dayan2001theoretical}.
Under this hypothesis, the instantaneous firing rate is sufficient information to predict the probabilities of spike sequences, and the spikes can be seen as generated by an inhomogeneous Poisson process.

We reduce the size of the time step $\Delta t$ until the probability that more than one spike could appear in $(t, t+\Delta t)$ is small enough to be ignored.
In this scenario, the neuronal response $y_i$ equals $1$ if there is a spike of neuron $i$, and $y_i=0$ if there is not.
The distribution $p(y_i\mid x_{1:t})$ is a Bernoulli distribution, and its relationship with the instantaneous firing rate $r_i(t)$ is
\begin{equation}\label{eq:500}\left\{
\begin{aligned}
&    p(y_i=0\mid x_{1:t})=1-r_i(t)\Delta t\\
&   p(y_i=1\mid x_{1:t})=r_i(t)\Delta t.
\end{aligned}\right.
\end{equation}
When the probability of a spike is small enough, according to \eqref{eq:450}, we have
\begin{equation}\label{eq:510}
    p(y_i=1\mid x_{1:t})= \alpha_i(x_t)p(y_i=1\mid x_{1:t-1}),
\end{equation}
where $\displaystyle\alpha_i(x_t)\equiv \frac{p(x_1=x_t\mid y_i=1)}{p(x_1=x_t\mid y_i=0)}$.
This probability may exceed the threshold that $\displaystyle p(y_i=1\mid x_{1:t}) \ll 1$ after a period of time, so in each time step, a squashing operation that keeps the probability small should be executed:
\begin{equation}\label{eq:515}
    p(y_i=1\mid x_{1:t})=\phi(p(y_i=1\mid x_{1:t})),
\end{equation}
where the function $\phi$ squashes the probability into the range $[r_{min}\Delta t, r_{max}\Delta t]$, and $r_{min}$ and $r_{max}$ are the maximum and minimum instantaneous firing rates, respectively.

We think that the optimization in \eqref{eq1.00} is performed in each time step, that is, for all $t\in N^+$, our goal is to
\begin{equation}\label{eq:518}
    \max_{p(y\mid x_{1:t})}I(x_{1:t};y).
\end{equation}
This is different from the scenario which is discussed in the last section, where the optimization is performed only in the time step $T$.
Compared to the previous case,
\eqref{eq:518} applies to the broader case where the latent variables can be slowly varying.
Given this, the second effect of the squashing operation in \eqref{eq:515} is to make the system placing more emphasis on recent stimuli and less on older ones.

Due to the property of temporal locality, the gradient information for learning is only available for the calculation in the latest time step, so according to \eqref{eq:510} 
\begin{equation}\label{eq:520}
    \delta p(y_i=1\mid x_{1:t})= p(y_i=1\mid x_{1:t-1})\delta \alpha_i(x_t).
\end{equation}
% \eqref{eq:520} is obtained under the premise that $p(y_i=1\mid x_{1:t-1})$ is small enough.
According to \eqref{eq3.30} and \eqref{eq:520},
\begin{equation}\label{eq:530}
    \delta I(x_{1:t};y)=\sum_{i=1}^{n}\E_{\begin{subarray}{l} x_{1:t}\sim p(x_{1:t})\\y_{-i}\sim p(y_{-i}\mid x_{1:t})\end{subarray}}p(y_i=1\mid x_{1:t-1})\log\frac{p(y_i=1\mid x_{1:t})}{p(y_i=1\mid y_{-i})} \delta \alpha_i(x_t).
\end{equation}
% \refstepcounter{equation}
From \eqref{eq:440}, \eqref{eq:520}, and \eqref{eq:530}, and similar to the derivation of \eqref{eq3.35}, the optimization problem \eqref{eq:518} can be solved with a distributed algorithm with the $i$'th part of it as
\begin{align}\label{eq:550}
 \asc_{\delta \alpha_i(x_t)}\E_{\begin{subarray}{l} x_{1:t}\sim p(x_{1:t})\\y_{-i}\sim p(y_{-i}\mid x_{1:t})\end{subarray}}p(y_i=1\mid x_{1:t-1})\log\frac{p(y_i=1\mid x_{1:t})}{p(y_i=1\mid y_{-i})} \delta \alpha_i(x_t).   
\end{align}
Similar to the derivation of \eqref{eq2.20}/\eqref{eq2.30} and \eqref{eq3.40}/\eqref{eq3.50}, \eqref{eq:550} can be solved by the chase of two algorithms \eqref{eq:560} and \eqref{eq:570}:
\begin{empheq}[left={\empheqlbrace}]{alignat=2}
    \des_{\delta q_i( y_{-i})}&\E_{\begin{subarray}{l} x_{1:t}\sim p(x_{1:t})\\y_{-i}\sim p(y_{-i}\mid x_{1:t})\end{subarray}} \left(q_i( y_{-i}) - p(y_i=1\mid x_{1:t})\right)\delta q_i( y_{-i})
    \label{eq:560}\\
    \asc_{\delta \alpha_i(x_t)}&\E_{\begin{subarray}{l} x_{1:t}\sim p(x_{1:t})\\y_{-i}\sim p(y_{-i}\mid x_{1:t})\end{subarray}}p(y_i=1\mid x_{1:t-1})\log\frac{p(y_i=1\mid x_{1:t})}{q_i( y_{-i})} \delta \alpha_i(x_t),\label{eq:570}
\end{empheq}
where $q_i( y_{-i})$ is the auxiliary function to fit the probability $p(y_i=1\mid  y_{-i})$.
Although \eqref{eq:570} seems complex at the first glance, it can be simply explained that the optimization is performed only when there is a spike in the previous time step, and the gradient signal of $\alpha_i(x_t)$ is proportional to the log-ratio of $ p(y_i=1\mid x_{1:t})$ and $ q_i( y_{-i})$.
% \begin{equation}
% \end{equation}
% \begin{equation}
%     \asc_{\delta \alpha_i(x_t)}\E_{\begin{subarray}{l} x_{1:t}\sim p(x_{1:t})\\y_{-i}\sim p(y_{-i}\mid x_{1:t})\end{subarray}}p(y_i=1\mid x_{1:t-1})\log\frac{p(y_i=1\mid x_{1:t})}{q_i( y_{-i})} \delta \alpha_i(x_t).\label{eq:570}
% \end{equation}

In summary, there are three operations to be performed in each time step for each representation neuron, including an update operation \eqref{eq:510}, a squash operation \eqref{eq:515}, and an optimize operation \eqref{eq:550}.
All of them use only information that is available locally in both space and time.

\paragraph{The squashing function $\phi$}
As mentioned above, there are two tasks for the function $\phi$. One is to keep the probability of spiking in a small value, and the other is to make the system forgetting earlier stimuli.
For the second purpose, it should be S-shaped and pass  through a fixed point at the prior probability $p(y_i=1)$.
The shape of $\phi$ may be not constant, but controlled by external factors such as attention(this topic will not be covered in this article).

The nonlinear region of $\phi$ continuously pushes the representation towards the fixed point (prior probability), which can account for the emergence of tuning curve \citep[see][chapter 1.2]{dayan2001theoretical}, since only the stimulus of specific patterns can resist this force well.

\paragraph{Continuous-time limit}
The system can approach continuous dynamical as the size of the time step $\Delta t$ shrinks to \SI{0}{\milli\second}.
In physical neural systems, there exists a refractory period of about \SI{1}{ms} after each action potential. 
However, we think of this phenomenon as a physiological defect rather than a necessary functionality.

\paragraph{Locality of operations}
All the three operations including \eqref{eq:510}, \eqref{eq:515}, and \eqref{eq:550} use only local information.
The remaining question is, as mentioned at the end of Section \ref{sc:3}, whether $q_i( y_{-i})$ and $\alpha_i(x_t)$ can be modeled with models without hidden units.

When $\Delta t$ is small enough, given that the probabilities $p(y_j\mid x_{1:t})|_{j\ne i}$ are small and independent, the concurrence of spikes from different neurons is negligible. In this case, the value of $q_i( y_{-i})$ will depend on $y_{-i}$ linearly as
\begin{equation}
    q_i( y_{-i})=(w_{(i)}^{\phantom{i}\top} y_{-i} + b_i)\Delta t,
\end{equation}
where $w_{(i)}$ is a trainable vector (synaptic strengths) of size $(n-1)$.

As for $\alpha_i(x_t)$, it is unreasonable to assume that the distribution of each component of $x_t$ is independent.
If $\alpha_i(x_t)$ is set to be linear, but we want it to handle complex problems like XOR, then this can be done by connecting multiple neurons hierarchically, each of which optimizes itself locally.
% 
% This may be one of the works to do in the future.
For example, in cat primary visual cortex, the responses of simple cells map linearly with light spots in their receptive fields, while the responses of complex cells, which receive input from other simple and complex cells, have complex and diverse mappings \citep{hubel1962receptive}.
However, this remains to be studied in future works.

\ifx\allfiles\undefined
\bibliography{reference}
\end{document}
\fi
\ifx\allfiles\undefined
\usepackage{my}
\begin{document}
\fi
% \section{Future works}

% % When we want the perceptual system to handle some complex problems, but 
% \paragraph{Short-term plasticity}
% We can introduce short-term plasticity to account for the workding memory and 
% \paragraph{Biological counterpart}

% \paragraph{Hierarchical structure}
\section{Summary}
In this article, under four assumptions, including
    (a) the Infomax principle,
    (b) the independent-neuron hypothesis,
    (c) the chronological order of sensory stimuli in a short time interval conveys no or little perceptible information, and
    (d) the independent spike hypothesis,
we obtain a biologically plausible learning algorithm with the proprieties of locality of operations, spike-based neural coding, and continuous-time dynamics.
The Infomax principle (a) is our cornerstone and starting point, and we have tried to convince readers that the assumptions (b), (c), and (d) are natural or necessary conditions for a neural perceptual system to acquire all the desired properties.
Given that (b), (c), and (d) have been postulated or discovered already in cognitive and physiological experiments, this enhances the plausibility that our algorithm can be found physically implemented in neural perceptual systems in nature.

% The backbone and originality of our argument rests on
% the idea that 

\ifx\allfiles\undefined
\bibliography{reference}
\begin{appendices}
\end{appendices}
\end{document}
\fi

% \section*{References}

\bibliography{mybibfile}
\newpage
\begin{appendices}
\section{Supplementary Information}
% app:1
\subsection{The proof of}
\label{app:1}
$\displaystyle\boxed{\delta I(x;y) = \E_{x\sim p(x)}\int \log(\frac{p(y\mid x)}{p(y)})  \delta p(y\mid x)\diff y}$
\begin{proof}
\mathleft
\begin{align}\label{eq:app10}
    \delta H(y) &=H_{p+\delta p}(y) - H_{p}(y)\nonumber\\
    &= \int-(p(y)+\delta p(y))\log (p(y)+\delta p(y)) \diff y-\int -p(y)\log p(y) \diff y\nonumber\\
    &= \int-(p(y)+\delta p(y))(\log p(y)+\frac{1}{p(y)}\delta p(y)) \diff y+\int p(y)\log p(y) \diff y\nonumber\\
    &=\int(-1-\log p(y))\delta p(y) \diff y\nonumber\\
    &= \int(-1-\log p(y))\int p(x)\delta p(y\mid x)\diff x \diff y\nonumber\\
    & = \E_{x\sim p(x)}\int(-1-\log p(y))\delta p(y\mid x) \diff y,
\end{align}
where $\delta p(y\mid x)$ represents the small change in $p(y\mid x)$, $\delta p(y)$ represents the small change in $p(y)$ due to $\delta p(y\mid x)$, $H_{p}(y)$ and $H_{p+\delta p}(y)$ represent the information entropies of $y$ when $y\sim p(y)$ and $y\sim (p(y)+\delta p(y))$ respectively.

\begin{align}\label{eq:app11}
    \delta H(y\mid x) =&H_{p+\delta p}(y\mid x) - H_{p}(y\mid x)\nonumber\\
    =& \E_{x\sim p(x)}\int-(p(y\mid x)+\delta p(y\mid x))\log (p(y\mid x)+\delta p(y\mid x)) \diff y-\nonumber\\
    &\E_{x\sim p(x)}\int -p(y\mid x)\log p(y\mid x) \diff y\nonumber\\
    =& \E_{x\sim p(x)}\int-(p(y\mid x)+\delta p(y\mid x))(\log p(y\mid x)+\frac{1}{p(y\mid x)}\delta p(y\mid x)) \diff y-\nonumber\\
    &\E_{x\sim p(x)}\int -p(y\mid x)\log p(y\mid x) \diff y\nonumber\\
    =& \E_{x\sim p(x)}\int(-1-\log p(y\mid x))\delta p(y\mid x) \diff y,
\end{align}
where $H_{p}(y\mid x)$ and $H_{p+\delta p}(y\mid x)$  respectively represent the conditional entropies when $y\sim p(y\mid x)$ and $y\sim (p(y\mid x)+\delta p(y\mid x))$ given $x\sim p(x)$.

\begin{align*}
    \therefore\delta I(x;y) =& \delta H(y) -\delta H(y\mid x)
    \\=& \E_{x\sim p(x)}\int \log(\frac{p(y\mid x)}{p(y)})  \delta p(y\mid x)\diff y
\end{align*}

\end{proof}

%app:1.2
\subsection{The proof of}
\label{app:1.2}
\noindent\fbox{%
    \parbox{\textwidth- 2\fboxsep-2\fboxrule}{%
The algorithm $\displaystyle\des_{\delta q(y)}\E_{x\sim p(x)} \int \left(q(y) - p(y\mid x)\right)\delta q(y)  \diff y
$ will end up with $q(y)=p(y)$.\hfill 
    }%
}%
\begin{proof}
\begin{align*}
    &\des_{\delta q(y)}\E_{x\sim p(x)} \int \left(q(y) - p(y\mid x)\right)\delta q(y) \diff y\\
    =&\des_{\delta q(y)}\int \left(q(y) - \E_{x\sim p(x)}p(y\mid x)\right)\delta q(y) \diff y\\
    =&\des_{\delta q(y)}\int \left(q(y) - p(y)\right)\delta q(y) \diff y.
\end{align*}
When $q(y) > p(y)$, this gradient descent algorithm will decrease $q(y)$, and vice versa. Therefore, this optimization will end up with $q(y) = p(y)$.
\end{proof}

\subsection{The proof of}\label{app:300}
\noindent\fbox{%
    \parbox{\textwidth- 2\fboxsep-2\fboxrule}{%
    For a family of distribution $p(a,b)$ without constraints on its marginal distributions, we can make a small change $\delta p(a\mid b)$, which is independent of $p(b)$, on $p(a\mid b)$ without changing $p(b\mid a)$ if and only if $p(a,b)=p(a)p(b)$.
\hfill 
    }%
}%
\begin{proof}\hfill\\
\mathleft
Only If:
When we make a small change $\delta p(a\mid b)$ to $p(a\mid b)$,
\begin{align*}
    \delta p(b\mid a)=&\frac{p(a,b)+\delta p(a\mid b)p(b)}{ p(a) + \int \delta p(a\mid b')p(b')\diff b'}-\frac{p(a,b)}{p(a)}\\
    =&\frac{\delta p(a\mid b)p(a)p(b)-p(a,b)\int \delta p(a\mid b')p(b')\diff b'}{[p(a)]^2}\\
    =&0
\end{align*}
\begin{align}\label{eq:a300}
    \therefore\frac{p(a)}{p(a\mid b)}=\frac{\int \delta p(a\mid b')p(b')\diff b'}{\delta p(a\mid b)}
\end{align}
$\because$ $\delta p(a\mid b')$ is independent of $p(b')$,\\
$\therefore$ we can choose different $p(b')$ without changing the equality of \eqref{eq:a300},\\
$\therefore$ $\delta p(a\mid b')$ is not a function of $b'$.\\
$\therefore$ $\displaystyle \frac{p(a)}{p(a\mid b)}=1$\\
$\therefore p(a,b)=p(a)p(b)$

\noindent If:\\
It is obviously true.
\end{proof}

\subsection{The proof of}
\label{app:310}
$\displaystyle \boxed{\delta I(x;y)=\sum_{i=1}^{n}\E_{\begin{subarray}{l} x\sim p(x)\\y_{-i}\sim p(y_{-i}\mid x)\end{subarray}}\int \log\frac{p(y_i\mid x)}{p(y_i\mid y_{-i})} \delta p(y_i\mid x) \diff y_i}$
\begin{proof}\ \\
\mathleft
According to \eqref{eq3.10},
\begin{align*}
    \delta p(y_i\mid xy_{-i})=\delta p(y_i\mid x)
\end{align*}
\begin{equation}\label{eq:a310}
\begin{aligned}
\therefore\delta p(y_i\mid y_{-i}) 
= &\int p(x\mid y_{-i})\delta p(y_i\mid xy_{-i}) \diff x\\
= &\int p(x\mid y_{-i})\delta p(y_i\mid x) \diff x.
\end{aligned}
\end{equation}
Let $\delta p_i(y) \equiv p(y_{-i})\delta p(y_i\mid y_{-i})$.
\begin{equation}\label{eq:a320}
\begin{aligned}
    \delta p(y) =&\int p(x)\delta p(y\mid x)\diff x\\
    =&\int p(x)\sum_{i=1}^{n}p(y_{-i}\mid x)\delta p(y_i\mid x)\diff x&&\text{(according to \eqref{eq3.20})}\\
    =&\sum_{i=1}^{n}\int p(y_{-i})p(x\mid y_{-i})\delta p(y_i\mid x)\diff x\\
    =& \sum_{i=1}^{n} p(y_{-i})\delta p(y_i\mid y_{-i}) 
    &&\text{(according to \eqref{eq:a310})}\\
    =&\sum_{i=1}^{n}\delta p_i(y)
\end{aligned}
\end{equation}
Let $H_{\delta p}(y)\equiv H_{p+\delta p}(y)-H_{p}(y)$. Since $\delta p$ and $\delta p_i$ are sufficiently small, according to \eqref{eq:a320},
\begin{equation}\label{eq:a330}
    \delta H(y)\equiv H_{\delta p}(y)=\sum_i^n H_{\delta p_i}(y).
\end{equation}
\begin{align*}
\because\int p(y)+\delta p_i(y) \diff y_i&=p(y_{-i})+\int \delta p_i(y) \diff y_i\\
&=p(y_{-i})+\int p(y_{-i})\delta p(y_i\mid y_{-i}) \diff y_i\\
&=p(y_{-i})
\end{align*}
\begin{equation*}
    \therefore H_{p+\delta p_i}(y_{-i})=H_{p}(y_{-i})
\end{equation*}
\begin{align*}
\therefore H_{\delta p_i}(y)&=H_{p+\delta p_i}(y)-H_p(y)\\
&=\left[H_{p+\delta p_i}(y_{-i}) + H_{p+\delta p_i}(y_i \mid y_{-i})\right]-\left[H_{p}(y_{-i}) + H_{p}(y_i \mid y_{-i})\right]\\
&=H_{p+\delta p_i}(y_i \mid y_{-i})-H_{p}(y_i \mid y_{-i})\\
&=\int p(y_{-i}) \int(-1-\log p(y_i\mid y_{-i}))\delta p(y_i\mid y_{-i}) \diff y_i\diff y_{-i}\\
&=\iint p(y_{-i}) (-1-\log p(y_i\mid y_{-i})) p(x\mid y_{-i})\delta p(y_i\mid x) \diff x\diff y_i\diff y_{-i}\\
&=\E_{\begin{subarray}{l} x\sim p(x)\\y_{-i}\sim p(y_{-i}\mid x)\end{subarray}}\int (-1-\log p(y_i\mid y_{-i})) \delta p(y_i\mid x) \diff y_i
\end{align*}
According to \eqref{eq:a330},
\begin{equation}\label{eq:a340}
    \delta H(y)=\sum_{i=1}^{n}\E_{\begin{subarray}{l} x\sim p(x)\\y_{-i}\sim p(y_{-i}\mid x)\end{subarray}}\int (-1-\log p(y_i\mid y_{-i})) \delta p(y_i\mid x) \diff y_i
\end{equation}
Let $\delta p_i(y\mid x)\equiv p(y_{-i}\mid x)\delta p(y_i\mid x)$. According to \eqref{eq3.20},
\begin{equation}\label{eq:a350}
    \delta p(y\mid x)=\sum_{i=1}^{n} \delta p_i(y\mid x)
\end{equation}
Let $H_{\delta p}(y\mid x)\equiv H_{p+\delta p}(y\mid x)-H_{p}(y\mid x)$. Since $ \delta p(y\mid x)$ and $\delta p_i(y_i\mid x)$ are sufficiently small, according to \eqref{eq:a350},
\begin{equation}\label{eq:a360}
    \delta H(y\mid x)\equiv H_{\delta p}(y\mid x)=\sum_{i=1}^{n} H_{\delta p_i}(y\mid x)
\end{equation}
According to \eqref{eq:app11},
\begin{align*}
    \delta H_{p_i}(y\mid x)=&\E_{x\sim p(x)}\int(-1-\log p(y\mid x))\delta p_i(y\mid x) \diff y\\
    =&\E_{x\sim p(x)}\int(-1-\log p(y_i\mid x)-\log p(y_{-i}\mid x))p(y_{-i}\mid x)\delta p(y_i\mid x) \diff y\\
    =&\E_{\begin{subarray}{l} x\sim p(x)\\y_{-i}\sim p(y_{-i}\mid x)\end{subarray}}\int(-1-\log p(y_i\mid x))\delta p(y_i\mid x) \diff y_i
\end{align*}
According to \eqref{eq:a360},
\begin{equation}\label{eq:a370}
    \delta H(y\mid x)=\sum_{i=1}^{n} \E_{\begin{subarray}{l} x\sim p(x)\\y_{-i}\sim p(y_{-i}\mid x)\end{subarray}}\int(-1-\log p(y_i\mid x))\delta p(y_i\mid x) \diff y_i
\end{equation}
According to \eqref{eq:a340} and \eqref{eq:a370}, 
\begin{align*}
    \delta I(x;y)=&\delta H(y)-\delta H(y\mid x) \\
    =&\sum_{i=1}^{n}\E_{\begin{subarray}{l} x\sim p(x)\\y_{-i}\sim p(y_{-i}\mid x)\end{subarray}}\int (-1-\log p(y_i\mid y_{-i}))\delta p(y_i\mid xy_{-i}) \diff y_i-\\
    &\sum_{i=1}^{n}\E_{\begin{subarray}{l} x\sim p(x)\\y_{-i}\sim p(y_{-i}\mid x)\end{subarray}} \int(-1-\log p(y_i\mid xy_{-i}))\delta p(y_i\mid xy_{-i}) \diff y_i\\
    =& \sum_{i=1}^{n}\E_{\begin{subarray}{l} x\sim p(x)\\y_{-i}\sim p(y_{-i}\mid x)\end{subarray}}\int \log\frac{p(y_i\mid xy_{-i})}{p(y_i\mid y_{-i})} \delta p(y_i\mid xy_{-i}) \diff y_i
\end{align*}
\end{proof}

\subsection{The proof of}
\label{app:3.5}
\noindent\fbox{%
    \parbox{\textwidth- 2\fboxsep-2\fboxrule}{%
The distributed algorithm $\displaystyle\asc_{\delta p(y_i\mid x)}\E_{\begin{subarray}{l} x\sim p(x)\\y_{-i}\sim p(y_{-i}\mid x)\end{subarray}}\int \log\frac{p(y_i\mid x)}{p(y_i\mid y_{-i})} \delta p(y_i\mid x) \diff y_i$, for $i\in (1:n)$, has the same stable point as $\displaystyle\asc_{\delta p(y\mid x)}I(x;y)$ when $\displaystyle p(y\mid x)=\prod_{i=1}^{n} p(y_i\mid x)$.\hfill 
    }%
}%
\begin{proof}\ \\
Let $\delta p_i(y\mid x)\equiv p(y_{-i}\mid x) \delta p(y_i\mid x)$, $\delta I_i(x;y)\equiv I_{\begin{subarray}{l} x\sim p(x)\\y\sim p(y\mid x)+\delta p_i(y\mid x)\end{subarray}}(x;y) - I(x;y)$, then
\begin{align*}
    \delta I_i(x;y)=E_{\begin{subarray}{l} x\sim p(x)\\y_{-i}\sim p(y_{-i}\mid x)\end{subarray}}\int \log\frac{p(y_i\mid x)}{p(y_i\mid y_{-i})} \delta p(y_i\mid x) \diff y_i,
\end{align*}
and so 
\begin{align*}
    \delta I(x;y)  = \sum_{i=1}^{n}\delta I_i(x;y).
\end{align*}
$\therefore$ If there exists a certain $\delta p(y\mid x)$ that makes $\delta I(x;y)>0$, then there must exist a $\delta p_i(x;y)$ that makes $\delta I_i(x;y)>0$. In return, if there exists a certain $\delta p_i(x;y)$ that makes $\delta I_i(x;y)>0$, then there must exist a $\delta p(y\mid x)=\delta p_i(x;y)$ that makes $\delta I(x;y)=\delta I_i(x;y)>0$.\\
$\therefore$ The distributed algorithm $\displaystyle\asc_{\delta p(y_i\mid x)}I_i(x;y)$, for $i\in (1:n)$, reaches the stable point as $\displaystyle\asc_{\delta p(y\mid x)}I(x;y)$.
\end{proof}

%app:3
\subsection{The proof of}
\label{app:3}
\noindent\fbox{%
    \parbox{\textwidth- 2\fboxsep-2\fboxrule}{%
The algorithm $\displaystyle\des_{\delta q(y_i\mid y_{-i})}\E_{\begin{subarray}{l} x\sim p(x)\\y_{-i}\sim p(y_{-i}\mid x)\end{subarray}} \int \left(q(y_i\mid y_{-i}) - p(y_i\mid x)\right)\delta q(y_i\mid y_{-i})  \diff y_i
$ will end up with $q(y_i\mid y_{-i})=p(y_i\mid y_{-i})$ when $\displaystyle p(y\mid x)=\prod_{i=1}^{n} p(y_i\mid x)$.\hfill 
    }%
}%
\begin{proof}\ \\
$\because$ $\displaystyle p(y\mid x)=\prod_{i=1}^{n} p(y_i\mid x)$\\
$\therefore$ $p(y_i\mid xy_{-i})=p(y_i\mid x)$
\begin{align*}
    &\des_{\delta q(y_i\mid y_{-i})}\E_{\begin{subarray}{l} x\sim p(x)\\y_{-i}\sim p(y_{-i}\mid x)\end{subarray}} \int \left(q(y_i\mid y_{-i}) - p(y_i\mid x)\right)\delta q(y_i\mid y_{-i})  \diff y_i\\
=&\des_{\delta q(y_i\mid y_{-i})} \iiint p(xy_{-i})\left(q(y_i\mid y_{-i}) - p(y_i\mid x)\right)\delta q(y_i\mid y_{-i})  \diff y_i\diff y_{-i} \diff x\\
=&\des_{\delta q(y_i\mid y_{-i})} \iiint p(y_{-i})\left(p(x\mid y_{-i})q(y_i\mid y_{-i}) - p(x\mid y_{-i})p(y_i\mid x)\right)\delta q(y_i\mid y_{-i})  \diff y_i\diff y_{-i} \diff x\\
=&\des_{\delta q(y_i\mid y_{-i})} \iiint p(y_{-i})\left(p(x\mid y_{-i})q(y_i\mid y_{-i}) - p(x\mid y_{-i})p(y_i\mid xy_{-i})\right)\delta q(y_i\mid y_{-i})  \diff y_i\diff y_{-i} \diff x\\
=&\des_{\delta q(y_i\mid y_{-i})} \iint p(y_{-i})\left(q(y_i\mid y_{-i}) - p(y_i\mid y_{-i}))\right)\delta q(y_i\mid y_{-i})  \diff y_i\diff y_{-i}\\
.
\end{align*}
When $q(y_i\mid y_{-i}) > p(y_i\mid y_{-i})$, this gradient descent algorithm will decrease $q(y_i\mid y_{-i})$, and vice versa. Therefore, this optimization will end up with $q(y_i\mid y_{-i}) = p(y_i\mid y_{-i})$.

\end{proof}

%app1.5

%app:4
\subsection{The proof of}\label{app:410}
\noindent\fbox{%
    \parbox{\textwidth- 2\fboxsep-2\fboxrule}{%
    \mathleft
If 
    \begin{equation*}\left\{
\begin{aligned}
   &p(y\mid x_{1:t}) \propto p(y)p(x_1\mid y)p(x_2\mid y)\cdots p(x_t\mid y)\\
   &p(x_{1}\mid y)\overset{d}{=}  p(x_{2}\mid y)\overset{d}{=}  \cdots  
\end{aligned}\right.
\end{equation*}
and
\begin{equation*}
    p(y\mid x_{1:t})=\prod_{i=1}^{n} p(y_i\mid x_{1:t}),
\end{equation*}
then
    \begin{equation*}
   p(y_i\mid x_{1:t}) \propto p(y_i)p(x_1=x_1\mid y_i)p(x_1=x_2\mid y_i)\cdots p(x_1=x_t\mid y_i)
\end{equation*}
\hfill 
    }%
}%

\begin{proof}\ \\
$\because$ $\displaystyle  p(y\mid x_1) =\prod_{i=1}^{n}p(y_i\mid x_1)\propto\prod_{i=1}^{n}p(y_i)p(x_1\mid y_i)=p(y)\prod_{i=1}^{n}p(x_1\mid y_i)$,\\ and\\
$\because$ $\displaystyle p(y\mid x_1)\propto p(y)p(x_1\mid y)$,\\
$\displaystyle \therefore p(x_1\mid y)=f(x_1)\prod_{i=1}^{n}g_i(x_1,y_i)$, where $f$ is a certain function, and $g_i(x_t,y_i)$ is a function that equals $ p(x_1=x_t\mid y_i)$. \\
$\because p(x_{1}\mid y)\overset{d}{=}  p(x_{2}\mid y)\overset{d}{=}  \cdots  $ \\
$\displaystyle \therefore p(x_t\mid y)=f(x_t)\prod_{i=1}^{n}g_i(x_t, y_i)$ \\
$\therefore$ $\displaystyle p(y\mid x_{1:t})\propto p(y\mid x_{1:t-1})\prod_{i=1}^{n}g_i(x_t, y_i)$\\
$\therefore$ $\displaystyle \prod_{i=1}^{n}p(y_i\mid x_{1:t})\propto \prod_{i=1}^{n}[g_i(x_t, y_i)p(y_i\mid x_{1:t-1})]$\\
$\therefore$ $\displaystyle p(y_i\mid x_{1:t})\propto g_i(x_t, y_i)p(y_i\mid x_{1:t-1})$
\mathleft
    \begin{equation*}\therefore
   p(y_i\mid x_{1:t}) \propto p(y_i)p(x_1=x_1\mid y_i)p(x_1=x_2\mid y_i)\cdots p(x_1=x_t\mid y_i)
\end{equation*}
\end{proof}

%app:5
% \subsection{}\label{app:5}
% To prove the prior probability $\displaystyle p_{y_i}=\frac{1}{1+(\frac{\epsilon_0^{\epsilon_0}(1-\epsilon_0)^{(1-{\epsilon_0})}}{\epsilon_1^{\epsilon_1}(1-\epsilon_1)^{(1-{\epsilon_1})}})^{\frac{1}{\epsilon_1-\epsilon_0}}} $.
% \begin{proof}\ \\
% The goal of the system is to $\displaystyle\max_{p(y\mid x)}I(x;y)$, and $\sum_{i}I(x;y_i)$ is an upper bound of $I(x;y)$.
% Ideally, $\displaystyle\max_{p(y\mid x)}I(x;y)$ will be reached when each $I(x;y_i)$ reaches its maximization and meanwhile $I(x;y)=\sum_{i}I(x;y_i)$\\
% Given $\epsilon_0\le p(y_i\mid x)\le \epsilon_1$,  in order to $\displaystyle\max_{p(y_i\mid x)}I(x;y_i)$, ideally, it should satisfy that\\
% $$
% \frac{\diff H(y_i)}{\diff p_{y_i}}
% =\frac{H_{p_x=\epsilon_1}(x)-H_{p_x=\epsilon_0}(x)}{\epsilon_1-\epsilon_0}
% $$
% $\therefore$
% $$
% \log\frac{1-p_{y_i}}{p_{y_i}}=\frac{[-\epsilon_1\log \epsilon_1 -(1-\epsilon_1)\log(1-\epsilon_1)]-[-\epsilon_0\log \epsilon_0 -(1-\epsilon_0)\log(1-\epsilon_0)]}{\epsilon_1-\epsilon_0}
% $$
% $\therefore$
% $$
% p_{y_i}=\frac{1}{1+(\frac{\epsilon_0^{\epsilon_0}(1-\epsilon_0)^{(1-{\epsilon_0})}}{\epsilon_1^{\epsilon_1}(1-\epsilon_1)^{(1-{\epsilon_1})}})^{\frac{1}{\epsilon_1-\epsilon_0}}}
% $$

% \end{proof}
\end{appendices}

\end{document}